\documentclass[runningheads]{llncs}
\usepackage{graphicx}
\usepackage{xcolor}
\usepackage{multirow}
\usepackage{float}
\usepackage{booktabs}
\begin{document}
\title{Improving Dribbling, Passing, and Marking Actions in Soccer Simulation 2D Games Using Machine Learning}
\titlerunning{Improving Dribbling, Passing, and Marking Actions in SS2D Games}
%
\author{Nader Zare\inst{1}\and Omid Amini\inst{4}\and Aref Sayareh\inst{5}\and Mahtab Sarvmaili\inst{1}\and Arad Firouzkouhi\inst{6}\and Stan Matwin\inst{1}\inst{2} \and Amilcar Soares\inst{3}}
\authorrunning{N. Zare et al.}
%
\institute{
Institute for Big Data Analytics, Dalhousie University, Halifax\\
\and
Institute for Computer Science, Polish Academy of Sciences, Warsaw\\
\and
Memorial University of Newfoundland, St. John's\\
\and
Qom University of Technology, Iran\\
\and
Shiraz University, Iran\\
\and
Amirkabir University of Technology, Iran\\
\email{\{nader.zare, mahtab.sarvmaili\}@dal.ca}\\ 
\email{amilcarsj@mun.ca}\\
\email{stan@cs.dal.ca}\\
\email{\{arefsayareh, omidamini360\}@gmail.com}\\
\email{arad.firouzkouhi@aut.ac.ir}\\
}
\maketitle              

\begin{abstract}
The RoboCup competition was started in 1997, and is known as the oldest RoboCup league. The RoboCup 2D Soccer Simulation League is a stochastic, partially observable soccer environment in which 24 autonomous agents play on two opposing teams. In this paper, we detail the main strategies and functionalities of CYRUS, the RoboCup 2021 2D Soccer Simulation League champions. 
The new functionalities presented and discussed in this work are (i) Multi Action Dribble, (ii) Pass Prediction and (iii) Marking Decision.
The Multi Action Dribbling strategy enabled CYRUS to succeed more often and to be safer when dribbling actions were performed during a game. 
The Pass Prediction enhanced our gameplay by predicting our teammate's passing behavior, anticipating and making our agents collaborate better towards scoring goals. 
Finally, the Marking Decision addressed the multi-agent matching problem to improve CYRUS defensive strategy by finding an optimal solution to mark opponents' players.


\vspace{2em}

\end{abstract}
\section{Introduction}

The idea of robotic soccer games was proposed as a novel research topic in 1992. Since then, the RoboCup has been considered the annual international competition for developing new ideas in A.I. and robotics.  
This competition is comprised of various leagues such as Rescue, Soccer Simulation, and Standard Platform leagues. 
Team CYRUS has participated in the annual RoboCup competitions and placed first, second, third, fourth, and fifth in RoboCup 2021, 2018, 2019, 2017, 2014. 
In RoboCup 2021, Cyrus played 21 games in total, winning nineteen games, and drawing two times.
CYRUS also won first place in the IranOpen in 2021, 2018, and 2014; first place in RoboCup Asia-Pacific 2018; and second place in the Japan Open 2020 competition. 


The rest of this paper is organized as follows. 
In Section 2, we present a new dribbling system and opponent behavior prediction.
Afterwards, we describe our new Pass Prediction module (Section 3) which is used to predict the action of a ball holder teammate. 
In Section 4, we detail the improvements of CYRUS' defensive strategy. 
Finally, we conclude our work and point to some directions for future works in Section 5.

\subsection{Previous Works}

Sixteen teams qualified for the 2D soccer simulation league in the 2021 RoboCup competition, including teams from Brazil, Canada, China, Germany, Iran, Japan, and Romania. 
In recent years, most of the teams have employed artificial intelligence algorithms to improve their game performance. 
For example, Helios has developed an algorithm called Player's MatchUp for exchanging players' positions \cite{hel21}.
FRA-UNIted has released a new 2D soccer simulation Python-based framework for performing reinforcement learning experiments \cite{fra21}.
ITAndroids optimized its field evaluator algorithm using Particle Swarm Optimization (PSO) and improved the goalkeeper performance for penalty kicks \cite{it21}.
Persepolis proposed an evolutionary algorithm to improve their offensive strategy \cite{per21} and YuShan applied Half Field Offense framework to build overall portraits of a team \cite{yu21}. 
CYRUS has concentrated its efforts on creating and applying machine learning techniques to improve its gameplay \cite{cyrus14,cyrus15,cyrus18,cyrus19}. 
In general terms, the improvements made in CYRUS are on the defensive decision-making method using Reinforcement Learning (RL), the opponents' behavioral analysis and prediction, and players' shooting skills.

\subsection{Release}

In this subsection we list several of our contributions towards increasing the popularity and improvement of 2D Soccer Simulation league competition. 

\subsubsection{Cyrus 2014 Source.}

As a part of our contribution to the development of the 2D Soccer Simulation league, we have released the Cyrus 2014 \cite{cyrus14} source code to encourage new teams to participate in the competitions. 
The source code can be found in github\footnote{Cyrus 2014 Source \url{https://github.com/naderzare/cyrus2014}.}.



\subsubsection{CppDNN.}
The C++ Deep Neural Network (CppDNN) library was developed by CYRUS team members to facilitate the implementation of Deep Neural Networks in the 2D Soccer Simulation environment. 
This library stores the weights of a neural network trained using the Keras library. 
The developed script within CppDNN transforms the trained weights of a deep neural network into a plain text file that is subsequently loaded to recreate the original deep neural network in C++.
In CYRUS, we use CppDNN to enhance our goalie performance, to predict opponent's movements against our dribbling agent, and to improve passing prediction between teammates.
The library can be found in our github\footnote{CppDNN Source Code \url{https://github.com/Cyrus2D/CppDNN}}.

\subsubsection{Pyrus - Python 2D Soccer Simulation Base.}
Most 2D soccer simulation teams exploit the Helios \cite{agent2d}, Gliders2d \cite{gldbase}, WrightEagle \cite{wrbase} or Oxsy \cite{oxsy} bases which are all developed in C++. 
Although those have shown fast processing and execution time, developing machine learning algorithms using C++ would be a time-consuming process.  
Due to the fast growth and popularity of the Python programming language among students and scientists and its plethora of libraries containing machine learning algorithms, the CYRUS team members have started developing an open-source python base for the 2D soccer simulation league. 
This base is currently available in the CYRUS github \footnote{Pyrus Base Source Code \url{https://github.com/Cyrus2D/Pyrus}} and it will support all features of the current 2D soccer simulation server in the Full-State mode in the near future.

\section{Multi-Action Dribble}
%
In soccer games, it is usual that the opponent's players tries to block the path of our ball holder, and a dribbling action can be helpful to escape from such situations. 
Dribbling also helps a player lead the ball forward and/or move it to a safer position.
Besides, some teams employ heavy defensive strategies that make the ball's movement extremely challenging. 
Therefore, dribbling is an essential skill to tackle harsh defensive strategies, and at the same time, to find a good spot for passing or shooting.  


Our team implemented an algorithm called Multi-Action Dribble (MAD) for improving our dribbling skills in SS2D games. 
MAD uses a Deep Neural Network (DNN) for predicting the opponents' movements so that the kickable player can find better positions to dribble. 
Before detailing MAD, we will first explain the agent 2D offensive algorithm known as \textbf{\emph{Chain Action}} and the \textbf{\emph{Basic Dribble}}. 
Once this background is given, we will detail how MAD works.

\subsection{Chain Action algorithm}

The Chain Action algorithm \cite{agent2d} employs the Breadth First Search (BFS) to make decisions for a kickable player. 
First, the Chain Action algorithm creates a decision tree with the root node of the game's current state.
Afterwards, simulated actions (e.g., Shoot, Pass, Dribble, etc) change the state of each node and create new children for this node. 
Then, a \textbf{\emph{Field Evaluator}} analyzes every node in the tree by the ball's predicted position. 
Finally, the best node is the node which has the maximum value of the evaluation and the action that leads the current state to this state is selected.


\subsection{Agent 2D Dribble Action Generator}

In the Agent2D base \cite{agent2d}, dribbling was developed in order to find a safe position for the kickable player to go without losing the ball or having an opponent intercept them. 
The Agent 2D Dribble Action Generator can only be evaluated on the game's current state (i.e., at the first level of the decision tree).
To evaluate the dribbling action, this agent first simulates turning in different directions (-180 to +180 with steps of 30 degrees) and dashing in each direction in order to create candidates for each position. 
In addition, for each point that the simulated agent has reached, the ball's velocity is calculated based on the cycles needed to reach that position so the agent can kick it before starting to dribble. 
The cycles required for each candidate position is detailed in Equation \ref{eq:dribble} and is equal to the sum of the number of turns and dashes plus one cycle for kicking the ball.

\begin{equation}
\label{eq:dribble}
    dribbleCycle = turnCycle + dashCycle + 1     
\end{equation}

\begin{figure}[ht]
    \centering
    \includegraphics[width=.55\textwidth]{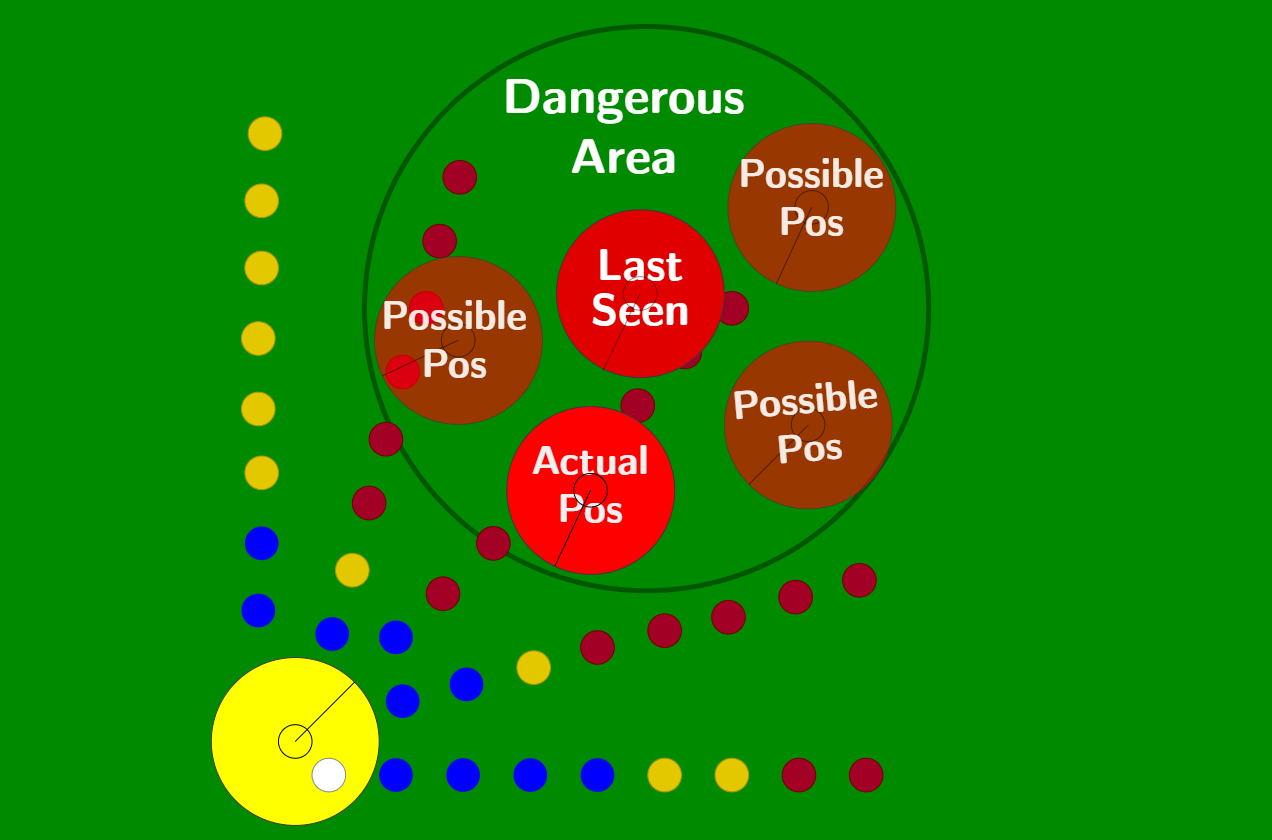} 
    \caption{Effecting pos-count on dribbling. Blue dots are possible candidates, red dots are candidates removed since they are impossible to be reach, and yellow dots are the possible candidates which are removed incorrectly because of the pos-count.}
    \label{fig:poscount}
\end{figure}

Next, for each player's dribbling candidate position, the opponent's players are evaluated to determine if they can reach the position before the agent or intercept the ball in the middle of the dribbling action. 
If any of these two situations occur, the candidate action is removed; otherwise, a new predicted state is created where the kickable player and the ball are now in the new position. 
Finally, a list of dribbles and predicted states are returned to the chain action tree.
Depending on the neck angle chosen by the agent, it might not have vision access to all of the field; therefore, the agent might not be able to see all opponents in every cycle. 
The number of cycles that an opponent has not seen is called their \textbf{\emph{pos-count}}. 
The agent should consider the pos-count and the last position of the opponent and an area is created which will likely contain the opponent. 
Therefore, a significant number of available candidates may be removed (Figure \ref{fig:poscount}) since there's always uncertainty in some areas not seen by the agent.

\subsection{MAD Generator}

Our team created a new dribble generator that inserts a simple action such as Kick, Dash, and Turn before the start of dribbling actions.
This generator makes a one-step action before the basic dribble generator, and its goal is to deceive the opponent to a wrong position so that the kickable agent may generate a better dribbling action. 
Therefore, the MAD generator only creates new child nodes from the current state (root of the tree).
As previously mentioned, the basic dribble generator only runs in the first layer. 
If the action of a parent state is generated by MAD, the basic dribble generator runs in the second layers as well.

\begin{figure}[ht]
    \centering
    \includegraphics[width=1.\textwidth]{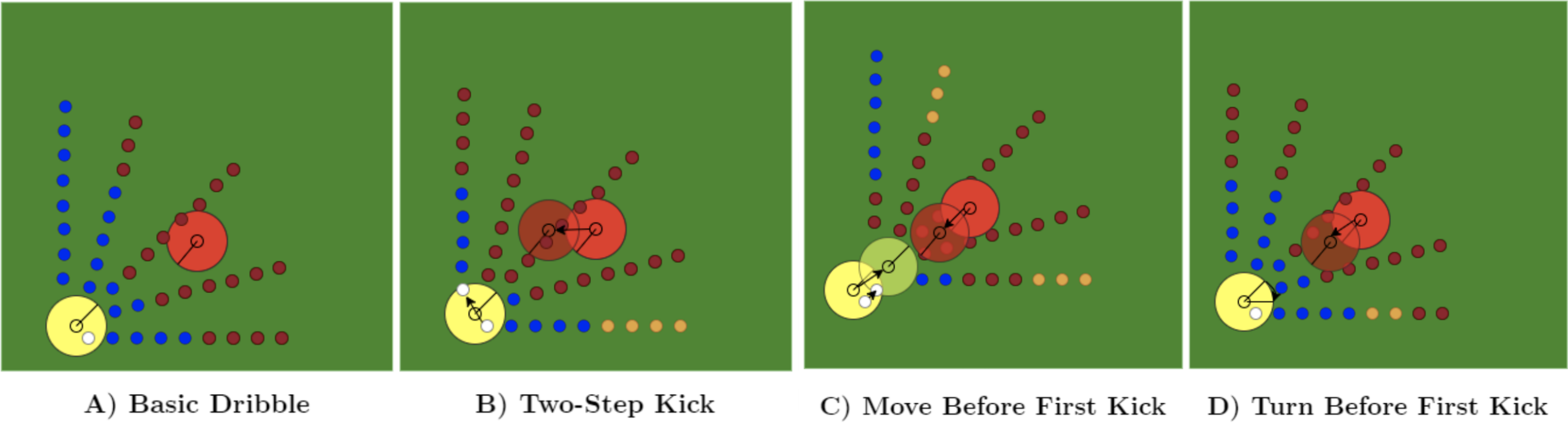}  
    \caption{Types of MAD and basic dribble, blue dots stands for possible actions generated by Basic Dribble, red dots show impossible dribbles, and orange dots demonstrate possible actions that have been added after using MAD.}
    \label{fig:madBasic}
\end{figure}

MAD generates three types of one-step action that are described below (Figure \ref{fig:madBasic}): 

\begin{itemize}
    \item \textbf{Two-Step Kick:} the agent kicks the ball so that it remains within the kickable area of the agent to distract opponent's player. This action changes the ball position and velocity.
    \item \textbf{Move Before First Kick:} The agent moves around the ball where the ball stays in the kickable area of the agent. This action changes the position of the player and updates the ball's position according to its velocity.
    \item \textbf{Turn Before First Kick:} The agent turns towards a direction where it causes some previous basic dribble candidates to become available after turning. However, it is important that the ball remains in the kickable area of the agent in next cycle. This action changes the direction of the player's body and updates the ball's position according to its velocity.
\end{itemize}




\subsection{Opponent movement prediction}

The Chain Action algorithm should predict the result of previous actions in the tree to find possible actions in the next layers after the first one. 
A predictor module in the Chain Action algorithm is available and is called State Predictor. 
The State Predictor should forecast the position and velocity of every object in the field after each action, but the implemented predictor in the Agent 2D base simply updates the position and velocity of the ball and the receiver teammate.
We added one cycle to the pos-count of opponents because the predictor was not forecasting the position of the closest opponents after MAD. 
Increasing the pos-count eliminates possible dribbling actions that might be considered difficult to perform. 
This problem led us to implement a position predictor module that receives information about an opponent and the ball, and then forecasts the position of the opponent after one cycle using a Deep Neural Network.

We collected data for training our DNN model running games between Cyrus and other teams. 
Each cycle was collected when a CYRUS player was kickable, and at least one opponent was near the ball (i.e., within 10 meters). 
The data set included the ball's position and the position of each opponent who was near the ball. 
All the positions are relative to the kickable agent's position and body direction. 
Next, we implemented a DNN using the position and velocity of the ball and the blocker, and the body direction of the blocker as input variables. 
The DNN architecture includes four hidden layers with $128$, $64$, $32$, and $16$ neurons, respectively. 
The output of this DNN is the opponent's predicted position, and its accuracy was $\pm 0.01$ meters.

\subsection{Results}

We executed some early tests with MAD, and MAD with the DNN predictor and verified their efficiency as follows. 
First, we trained DNNs to predict the opponent's position by running $1000$ games against YuShan and Persepolis. 
The results are shown in Table \ref{table:Table0} where we present CYRUS's winning probability, without MAD and the DNN Predictor, with MAD but without the DNN Predictor, and with both when playing against Yushan and Persepolis.
The results show that our winning rate against both teams increases when using both MAD and the DNN predictor when playing against both teams.
One interesting result is that using MAD alone actually decreases our winning rate (we believe this is due to the use of the additional pos-count alone) against both teams. Once the DNN predictor is included, we see that the rate increases. 


\begin{table}[ht]
    \centering
    \caption{Winning rate of CYRUS using MAD and DNN}
    \label{table:Table0}
    \begin{tabular}{@{}lcc@{}}
    \toprule
    \textbf{Experiments}             & \textbf{Yushan} & \textbf{Persepolis} \\ \midrule
    CYRUS                            & 77\%            & 88\%                \\
    CYRUS with MAD                   & 76\%            & 85\%                \\
    CYRUS with MAD and DNN Predictor & 80\%            & 90\%                \\ \bottomrule
    \end{tabular}
\end{table}
\section{Pass Prediction}

As in real soccer, passing behavior in a 2D Soccer Simulation (SS2D) game plays a critical role in increasing the chance of winning. 
A team with excellent passing skills prevents opponents from scoring, may create better chances of scoring themselves, and may conserve stamina. 
We believe that strengthening a team's passing decision-making algorithm will lead the team to have better performance and to win games.
However, the random noises from the environment in the partial observation of the agents is a major challenge the players face while choosing their actions since it creates uncertainty on the best action to take in a given moment. 
Many approaches such as Monte Carlo or Kalman Filter were used in the past to address this problem, but in CYRUS we took a different route. 
In our work, we attempted to predict the action of the ball owner if our agent had pure observation data (i.e., no noise generated by the simulator). 
We used a full state action predictor from noisy observation that is detailed in \cite{cyrus21} and features engineered in \cite{cypaper} to improve the prediction of the behavior of our ball owner player and are detailed below. 

In \cite{cyrus21}, the full state action predictor from noisy observation is trained to receive noisy observations from the server and to forecast the action of a player if it gets an observation without noise.
The soccer simulation server has another option known as \textbf{\emph{full-state mode}} which does not apply the random partial noises observations to the agents' vision. 
When this option is enabled, the server also sends normal observation.
We have also developed a module named \emph{Data Extractor} to create training data to feed machine learning models in an SS2D game (more details are available in \cite{cyrus21} and \cite{cypaper}).
In summary, this module collects the events of the game and transforms these events into training data for machine learning models.

\subsection{Experimental Setup}

For the purposes of evaluating passing actions we used the data extractor module as follows. 
The data extractor module generates a new data instance for each cycle that one of our agents is the ball holder and the selected action is a pass.
For the early results showed in \cite{cyrus21}, we generated $794$ features from these observations, but for playing in the Robocup 2021, we used the features presented in \cite{cypaper}.
In \cite{cypaper}, each data instance contains 12 features for the ball, 42 for each one of our players ($42 \times 11$), and 24 for each of the opponent's players ($24 \times 11$), totaling $738$ features. 
The list of extracted features is divided into nine feature subgroups that are measured for the ball, our players, or the opponents’ players. 
These nine groups are Position, Kicker, Velocity, Body, Team, Player Type, Top k high-risk opponents, Top k nearest opponents, and Goal. 
Since at each time step, the ball holder is responsible for generating a data instance, the module creates all of these features for all agents in the field.


We used two labels for each data instance from the sorting module proposed in \cite{cypaper} which are the Index Number and Uniform Number (Unum). 
The Uniform Number is the unique number of a target agent (who may receive the ball) in the game. 
Differently from Unum, the Index Number refers to the index of a target agent in the sorted data. 
The sorting module is an essential component in the data preparation step since it organizes the input features and creates the training data set for our machine learning model.
We have also used the \textit{Kicker be First} field which is a binary attribute that pushes the features of the ball holder as the first element of data \cite{cypaper}. 
After sorting the data, this field assigns a label to them that is based on the index of ball receiver. 
Therefore, applying two sorting methods and changing the \textit{Kicker be First} attribute (i.e., true or false), we generated four different experiments with a different order of the input data. 

\vspace{-3mm}
\subsection{Result} 
Differently from what was presented in \cite{cypaper}, where we showed the effects of the $738$ features when forecasting the player's pass action using Agent2D as the base team, we show in this paper the results of using CYRUS as our base team.  
In Table \ref{tbl:table2}, we compared the features used in \cite{cyrus21} and the features used in \cite{cypaper} with (Noisy) and without (Pure) the presence of noise when running matches. 
We ran over $1000$ games for collecting data and testing our methods using CYRUS as the base team and the results are presented in Table \ref{tbl:table2}. 
When using the data without noise, we see that the features presented in \cite{cypaper} when combined with the proposed strategies produces pass prediction accuracy rates ranging from $76.23$ to $80.51$. 
Although these results are very promising, this setup using data without noise is not the one used in the Robocup competitions. 
Therefore, we tested the differences between the features shown in \cite{cyrus21} and \cite{cypaper} using the proposed strategies when noise was present. 
The results in Table \ref{tbl:table2} shows that the $738$ features of \cite{cypaper} produces generally better accuracy values (e.g., a difference ranging around 8\% to 10\%) when compared to the $794$ features of \cite{cyrus21}.

\begin{table}[ht]
\centering
\caption{Accuracy of the two models for six datasets and three feature groups}
\label{tbl:table2}
\begin{tabular}{@{}ccccc@{}}

\toprule
\multicolumn{2}{c}{\textbf{Strategies}}       & \multicolumn{3}{c}{\textbf{Features}}                                                                                                                                                                                                                  \\ 
\textbf{Sorting}   & \textbf{Kicker be first} & \textbf{\begin{tabular}[c]{@{}c@{}}Features \cite{cypaper} \\ Pure\end{tabular}} & \textbf{\begin{tabular}[c]{@{}c@{}}Features \cite{cyrus21}\\ Noisy \end{tabular}} & \textbf{\begin{tabular}[c]{@{}c@{}}Features\cite{cypaper}\\ Noisy\end{tabular}} \\ \midrule
Uniform   & False           &      80.51 &      57.90  & 67.86                                                                \\
Uniform & True            &       79.60      &    58.74  &    67.63                                                             \\
X-Sorting   & False           & 76.23     &     53.64 &     63.86                                                            \\
X-Sorting & True            &   77.48     & 57.60     &     65.13                                                            \\ \bottomrule
\end{tabular}
\end{table}
\vspace{-5mm}



\section{Improved Marking Decision}

In a soccer game, marking is a defensive strategy that helps to prevent an opposing team member from taking control of the ball. 
Several marking strategies exist in soccer such as \textit{man-marking} and \textit{zonal-marking}.
In the man-marking strategy, defenders have to mark a specific opponent player, and zonal-marking is a defensive strategy where defenders have to cover an area of the field \cite{soccer_book}.
Defining a marking strategy is one of the most challenging actions in defensive algorithms for soccer 2D games. 
The challenge is mainly to be able to synchronize the agents' decisions, despite different and noisy partial observational views.

In this section we will first discuss some marking strategy algorithms that are based on greedy strategies. 
Then we will explain how the CYRUS team strives to minimize the effects of the challenges imposed by greedy strategies by using strategies called \textbf{\emph{"Optimized multi-agent matching" (OMAM) }} and \textbf{\emph{Player Grouping}} method.

\subsection{Marking Algorithms}
\subsubsection{Proximity-Based Marking.}
Using Proximity-Based Marking, each player simply selects their nearest opponent to mark. 
This algorithm is one of the most rudimentary solutions and has several problems. For example, it is common when using this strategy that more than one player decides to mark a single opponent, and as a consequence, several opponents are left unmarked.

\subsubsection{Danger-Based Marking}
\begin{figure}[ht]
    \centering
    \includegraphics[width=.5\textwidth]{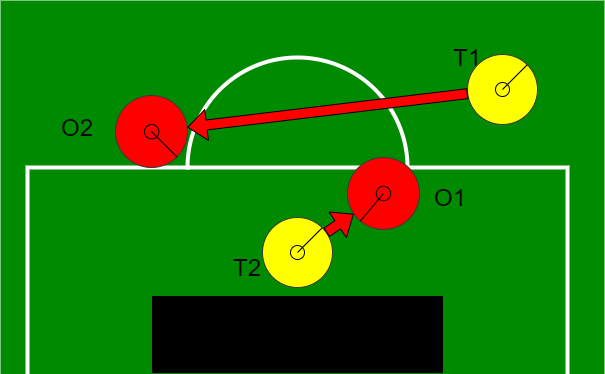} 
    \caption{Improper defense strategy, leaving the leftmost player empty for a longer time, while T1 should mark O1 and T2 should mark O2 for an optimal solution.}
    \label{fig:greedy_mark}
\end{figure}

A solution to solve the main problem of \emph{Proximity-Based Marking} which we explored was to rank opponents based on how dangerous they are using various attributes such as distance to our goal and distance to the ball. 
After sorting them, we assign the closest player who is not marking anyone to the most dangerous players. More specifically, the first most dangerous opponent is marked by the closest teammate, and the second opponent is marked by the closest teammate, which is not marking the most dangerous opponent, and so on.
This algorithm, while theoretically may avoid multiple players marking a single opponent fails when each player is computing the dangerous opponents and distances separately and each player has different and noisy observational views of the game. 
An example is shown in Figure \ref{fig:greedy_mark}, where the greediness of this algorithm stops it from reaching a solution that is good for the team as a whole.

 
\vspace{-5mm}
\subsubsection{The Hungarian Method.}

To find an optimal solution and eradicate the issues found with acting in a greedy manner such as the one presented in Figure \ref{fig:greedy_mark}, each agent must also consider the difficulty of other agents marking each opponent.
The Hungarian Matching algorithm receives as input the number of agents, tasks which must be completed, and the cost for each agent to do each task as input. 
After, it assigns exactly one agent to each task, in a way that minimizes the total cost of completing all of the tasks. 

We used this method in the 2D soccer simulation league by assigning the distance between agents and their opponents as their costs. The result is the minimal total movement of the team towards covering all of the opponents.
However, this method does not improve the synchronization problem that arises from our agents having different observations. 
In fact, the problem is worsened as our solutions become more dependant on observation and require more information about other players' positions.
As there are $10$ factorial ways that an entire team can mark their opponents (not including goalkeepers), even a little noise can disrupt the synchronization of the agents. This results in our agents calculating different solutions, which would lead to two or more agents marking a single opponent and leaving some opponents unmarked.
The number of agents and the number of opponents that need to be covered are not always the same, as the opposing defense does not need to be covered. Also, our offensive players should not mark any players to be ready for a counter-attack and to avoid wasting stamina.

In summary, using Hungarian method has the following problems: (1) the number of tasks and agents should be equal; (2) the method does not consider the importance of each task; and (3) is very susceptible to observational noise, as the number of candidate solutions is high.

\subsubsection{Optimized Multi-Agent Matching (OMAM).} 
The new method we used in Robocup 2021 aims to avoid the first and second problems of the \emph{Hungarian Method} by combining ideas from Danger-Based Marking and Multi-Agent Matching. 
This method handles synchronization by decreasing the total possible candidate solutions that are considered using K-top tasks.

This algorithm contains multiple steps to find the best solution for assigning agents' tasks. 
In OMAM, a solution is a collection which includes pairs of agents and tasks to be performed. 
The number of pairs are less than or equal to the minimum number of tasks or agents. 
To compare two solutions, we assign two values to each solution. 
The first value is the summation of the pairs' cost in the solution. 
To calculate the second value, we sort the tasks by using their importance value, then we generate a string containing 0 and 1 (i.e., if a task has been assigned in the solution we add 1 for the task, otherwise we add 0). 
A solution A is better than B if its second value is greater than B's. 
If their second values are the same, the solution with a lower first value is taken. 
To decrease time complexity to find the best solution, we keep pairs in the best k-tasks for each agent. 
In CYRUS, the value of k is set to be three, meaning we remove all tasks except the best three ones for each agent.
Any solution A with an opponent with a higher value marked will have a higher second score than any other solution such as B. 
We first process solutions that have the more dangerous opponents covered since it will have a higher second value when compared with any other solutions that only have less dangerous opponents. 
We use this idea to optimize the time complexity of this algorithm since we do not need to search for any other candidate with a lower score.
Keeping k-tasks helps our players to improve the synchronization of the agents and time complexity.
To further decrease the number of solutions, we separated our players into three groups: Back, Middle, and Forward players. 
We also divided opponents players into two groups: Attacker and Normal.
In the CYRUS defense algorithm, first, our Back players mark the opponent's Attackers. Then our Middle players mark unmarked Attackers' opponents. 
Finally, our free players, including Middle and Forwards except for Back players, try to mark other unmarked opponent's players. 
Therefore, the number of solutions decreases to $6$ factorial.

\subsection{Results}
The effectiveness of OMAM can be seen comparing the average goals our team has conceded in the main round over $28$ games in 2019 and 2021. 
OMAM has improved the Cyrus defense strategy from an average of $0.9$ goals taken in main round in 2019 competition to $0.33$ in 2021. We do believe that the opponents' offenses have improved as well within these 3 years so the results are interpreted as even better from our perspective. 

\section{Conclusion}
In this paper we detailed the major algorithms and innovations of our team CYRUS that we believe led us to become the RoboCup 2021 champions in the Soccer Simulation League 2D.
In the past two years, and in in-between competitions, we improved many algorithms in CYRUS and saw drastic improvements in our positions in competitions. 
In the future, we will release the first official version of PYRUS 2D base which is written in Python 3 and open several venues to apply machine learning techniques in Soccer Simulation League 2D. 

\end{document}